\documentclass[conference]{IEEEtran}
\IEEEoverridecommandlockouts

\usepackage{cite}
\usepackage{amsmath,amssymb,amsfonts}
\usepackage{algorithmic}
\usepackage{graphicx}
\usepackage{textcomp}
\usepackage{xcolor}
\def\BibTeX{{\rm B\kern-.05em{\sc i\kern-.025em b}\kern-.08em
    T\kern-.1667em\lower.7ex\hbox{E}\kern-.125emX}}

\usepackage{helpers/style}

\begin{document}

\maketitle

\begin{abstract}
    Anomaly detection is inherently characterised by severe class imbalance, making the interpretation of evaluation metrics challenging.
Although metrics such as AUROC, AUPR, F$_1$-score, and MCC are widely used, their values convey different meanings depending on the anomaly ratio.
In this work, we analyse the behaviour of those four common anomaly detection metrics under varying levels of imbalance.
We focus on the study of \textit{metric landscapes}, visualisations that relate metric values to true positive and true negative rates, providing an intuitive view of metric preferences and stability.
Our analysis offers practical guidance for interpreting and comparing anomaly detection results across datasets with different imbalance ratios.

\end{abstract}


\section{Introduction}

Evaluation metrics are central to machine learning, providing compact summaries of model performance that enable comparison across methods, datasets, and experimental settings.
They are essential in practice due to the scale of modern benchmarking, where exhaustive qualitative analysis is intractable.
Metrics also support model selection under different application priorities, such as balancing detection performance against false positives.
However, reducing model behaviour to a single scalar inevitably discards information, and the interpretation of these scores is often non-trivial.

\begin{figure}[t]
    \centering
    \includegraphics[width=\linewidth]{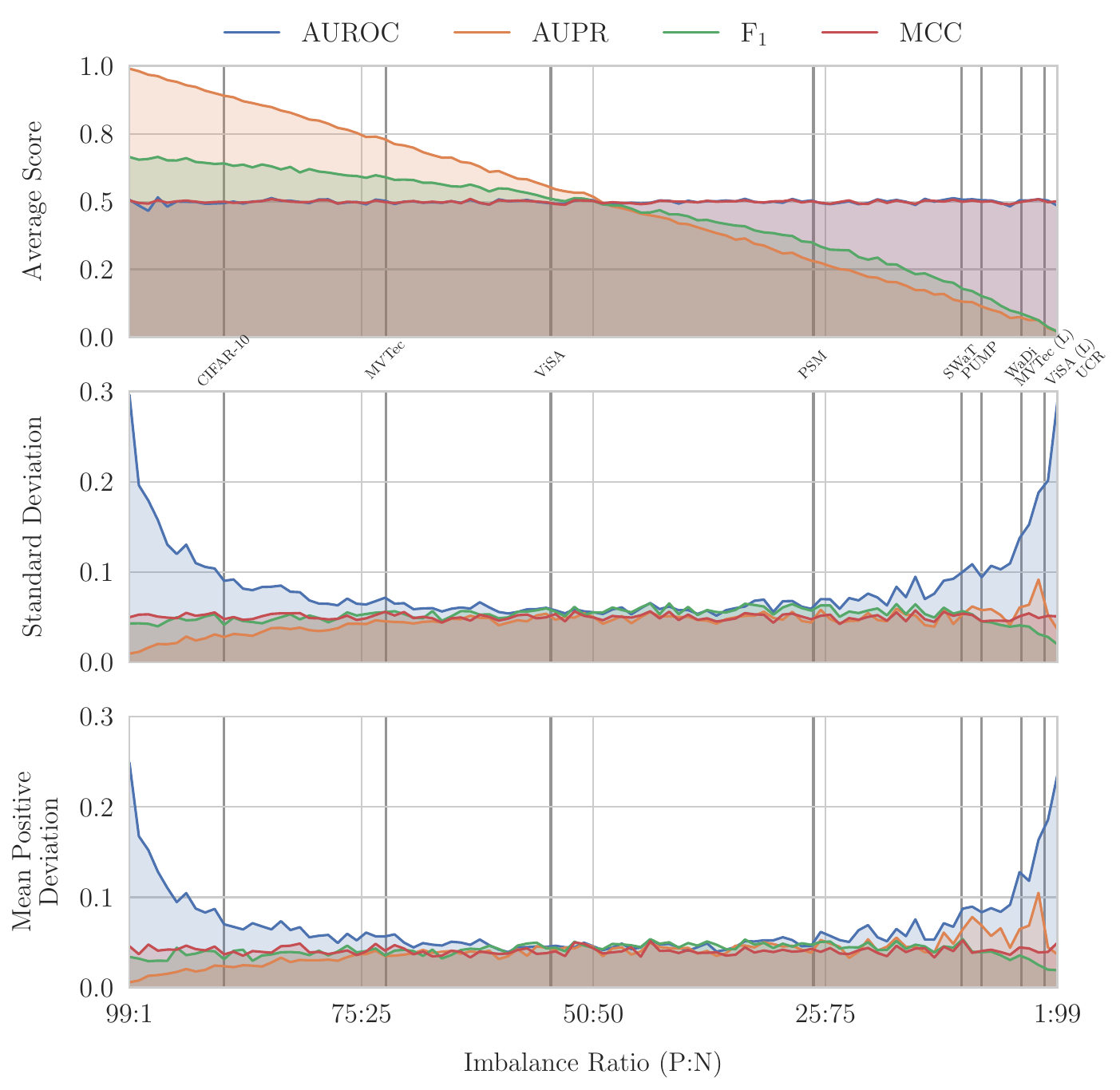}\\[-.5em]
    \caption{
Random baselines are evaluated across different target classes balance.
For each setting, metrics are computed over 100 runs with random predictions.
We report the mean, standard deviation, and mean positive deviation, noting that approximately half of the deviations being positive per metric.
}
    \label{fig:stability}
\end{figure}

\begin{figure*}[t]
    \centering
    \includegraphics[width=\linewidth]{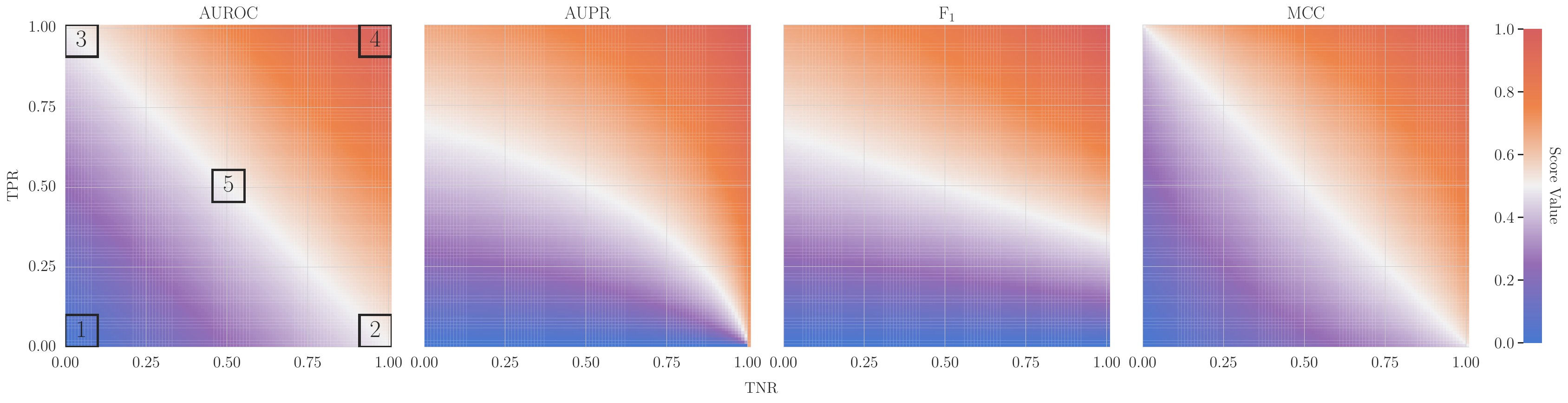}\\[-.5em]
    \caption{
Metrics landscapes in the balanced case (50:50).
Colour follows scores \wrt{} True Positive Rate (TPR) and True Negative Rate (TNR).
}
    \label{fig:landscapes}
\end{figure*}

This issue is particularly pronounced in anomaly detection, where the data is inherently and extremely imbalanced.
In many practical settings, training is unsupervised and performed only on normal data, which prevents any direct mitigation of class imbalance during learning.
Evaluation is then conducted under highly skewed test distributions with rare anomalies, making performance assessment delicate~\cite{Mejri2022UnsupervisedAD}.

To address class imbalance, weighted or balanced variants of standard measures are commonly used~\cite{Reinke2021CommonLO}, but they are generally avoided in anomaly detection as they decouple evaluation from the operational class distribution.
Instead, anomaly detection primarily relies on metrics less sensitive to class proportions, notably AUROC, as well as AUPR, F$_1$-score, and MCC which are less affected by negative class dominance~\cite{Hossain2024EvaluatingAD}.
Despite known limitations under severe imbalance and proposed alternatives, no consensus has emerged and these standard metrics continue to dominate evaluation.

We argue that these metrics remain useful provided their interpretation accounts for how their values depend on both prediction quality and class imbalance, since identical scores can correspond to different classifier behaviours under varying anomaly ratios.
This work therefore study how metric values evolve as functions of True Positive Rate (TPR, or Recall, Sensitivity) and True Negative Rate (TNR, or Specificity) across different levels of anomaly ratios.
To do so, we utilise \textit{metric landscapes}, which map each metric score over the space of sensitivity–specificity operating points.
This framework allows us to visualise how different metrics emphasise different trade-offs between sensitivity and specificity and how these preferences shift under increasing imbalance.

In summary, this work provides a systematic analysis of four widely used metrics in anomaly detection (AUROC, AUPR, F$_1$, and MCC), studying their behaviour across a wide range of class imbalance ratios.
We introduce metric landscapes as an intuitive framework for analysing metric values as functions of TPR, TNR, and class imbalance.
We focus on settings representative of image and time-series anomaly detection benchmarks, as well as one-class classification, providing reference points for interpreting metric values and understanding their behaviour under realistic imbalance conditions.

\section{Related Work}

Research on evaluation under class imbalance has focused on identifying the unreliability or suitability of metrics when class distributions are skewed, and determining the best ones for specific applications.
Existing literature analyses the theoretical and empirical properties of commonly used metrics such as Accuracy, F$_1$, AUROC, AUPR, MCC, Cohen’s Kappa, and Geometric Mean under varying imbalance ratios \cite{Chicco2020TheAO,Luque2019TheIO,delaCruzHuayanay2024PerformanceOE,Halimu2019EmpiricalCO,Fatourechi2008ComparisonOE}.
These studies consistently highlight that several widely used measures, including Accuracy and F$_1$, can become misleading under strong imbalance, while metrics such as MCC or AUROC are often more stable.
Additional work has further investigated relationships between metrics and the redundancy of the information they convey \cite{Walauskis2025ChoosingTR}.

Some studies focus on domain-specific evaluation, aiming to identify metrics that best reflect model prediction profiles.
Evaluation is often tied to comparing classifiers trained with different sampling, weighting, or thresholding strategies \cite{Chennuru2011OnTC,Abdelhamid2024BalancingTS,Kumar2016OnTC,Hasanin2019ACO,Hancock2023EvaluatingCP}.
These analyses are typically grounded in application areas such as healthcare, finance, fraud detection, and network security, and seek to determine which metrics best align with desired model behaviours across different datasets.
In parallel, other works propose new metrics or evaluation frameworks tailored to specific domains, particularly in time-series anomaly detection, where temporal structure motivates alternative formulations of performance~\cite{Kovcs2019EvaluationMF,Kim2022ASO,Srb2023NavigatingTM}.
These approaches are motivated by the view that no single metric can fully capture all relevant aspects or characterise underlying model behaviour and performance.

\begin{figure*}[t]
    \centering
    \includegraphics[width=\linewidth]{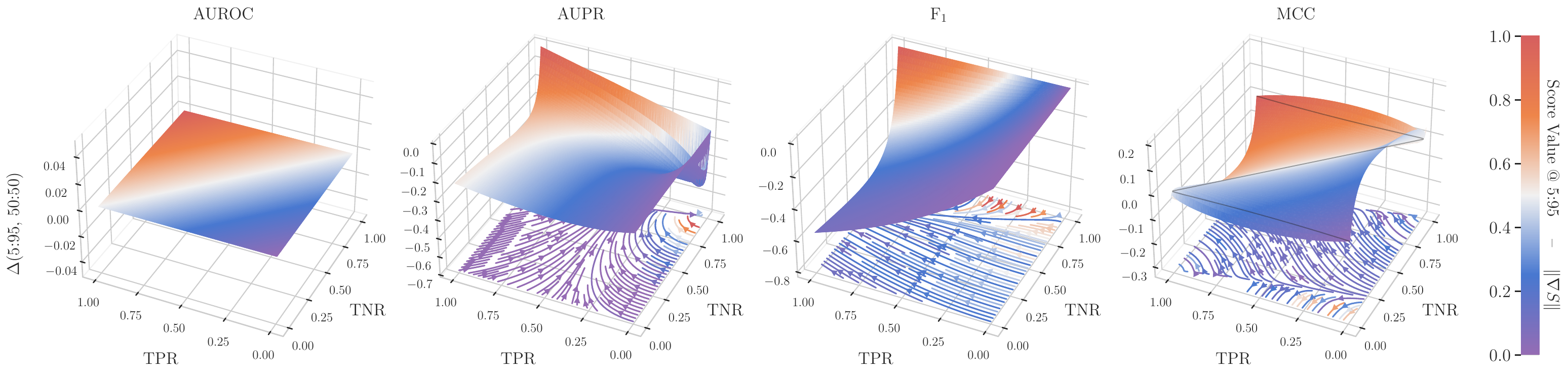}\\[-.5em]
    \caption{
Landscape evolution between the balanced (50:50) and negative-majority settings (5:95) shown as the landscapes difference.
Surface colour follows the scores at the 5:95 ratio.
The 0-level line intersection is displayed in black.
Surface gradient directions are projected onto the figure bottom plane to improve terrain understanding, and coloured according to the normalised gradient magnitude.
Gradient arrows point downhill.
$\Delta(a,b)=a-b$.
}
\label{fig:evol-neg}
\end{figure*}

\begin{table}[t]
    \centering
    \caption{Summary of datasets categories and statistics.}
    \label{tab:datasets}
    \resizebox{.9\linewidth}{!}{
\renewcommand{\arraystretch}{1.15}%
\begin{tabular}{|c|c|c|c|}
\hline\textbf{Dataset} & \textbf{Type} &\textbf{Test Set Size} &\textbf{Anomaly \%}\\
\hline
\hline
UCR~\cite{Wu2023CurrentTS}&UTS&14,050,817&00.19\\

{WaDi}~\cite{Ahmed2017WADIAW}  & MTS  & {172,801} & {05.77}  \\
{PUMP}~\cite{Feng2021TimeSA} & MTS & {143,401} &  {10.05} \\
{SWaT}~\cite{Mathur2016SWaTAW} & MTS  & {449,919} &  {12.14} \\
{PSM}~\cite{Abdulaal2021PracticalAT}& MTS & {87,841} & {27.76} \\

\hline
CIFAR-10~\cite{Krizhevsky2009LearningML}&OCC&10,000& 90.00\\
\hline
VisA~\cite{Zou2022SPottheDifferenceSP}&IAD&2,162&55.50\\
VisA~\cite{Zou2022SPottheDifferenceSP}&IAD (L)&566,755,328&00.61\\
MVTec~\cite{Bergmann2019MVTecA}&IAD&1,725&72.93\\
MVTec~\cite{Bergmann2019MVTecA}&IAD (L)&452,198,400&03.33\\
\hline
\end{tabular}
}
\end{table}

We complement existing studies by investigating how the meaning of a given metric score changes across imbalance settings and, more generally, how commonly used evaluation metrics in anomaly detection respond to varying levels of class imbalance.
We ground our analysis in anomaly ratios that are representative of real image, one-class, and time-series anomaly detection datasets, while explicitly avoiding the goal of identifying a best metric for a given scenario or determining which metric best reflects a particular model behaviour.

\section{Preliminaries}

This section presents the anomaly detection applications considered in this work, along with the datasets used to derive the anomaly ratios supporting the study (\Cref{subsec:datasets}).
It then introduces the evaluation metrics analysed under these varying imbalance regimes (\Cref{subsec:metrics}).

\subsection{Applications \& Datasets}
\label{subsec:datasets}

We consider three main applications of anomaly detection: Time-Series Anomaly Detection (TSAD), One-Class Classification (OCC), and Image Anomaly Detection (IAD).
These settings differ in the nature of the data, its structure, and primarily in anomaly granularity and the imbalance induced by the evaluation protocol.

\subsubsection{Time-Series Anomaly Detection (TSAD)}
TSAD covers sequential observations indexed in time, typically acquired from one or more sensors at regular sampling intervals.
The objective is to identify anomalous temporal segments, \ie, contiguous intervals in which the observed dynamics deviate from normal behaviour~\cite{Hermary2026ASTERLP}.
Anomalies may appear as irregular values, ranging from short-lived extreme deviations to longer, context-dependent patterns, and are rare by nature.

We distinguish between Univariate Time-Series (UTS), involving a single observation channel, and Multivariate Time-Series (MTS), involving multiple synchronised sensor streams.
In the UTS setting, we consider the UCR anomaly archive~\cite{Wu2023CurrentTS}, which contains 250 heterogeneous time-series drawn from domains such as healthcare and finance, each with one anomalous segment.
In the MTS setting, we consider four benchmark datasets with varying anomaly ratios: Pooled Server Metrics (PSM)~\cite{Abdulaal2021PracticalAT}, PUMP Fillage (PUMP)~\cite{Feng2021TimeSA}, Secure Water Treatment (SWaT)~\cite{Mathur2016SWaTAW}, and Water Distribution (WaDi)~\cite{Ahmed2017WADIAW}.
These datasets are continuous multivariate signals with multiple labelled anomalous intervals.

\subsubsection{One-Class Classification (OCC)}
The objective in OCC is deriving a boundary around an observed (normal) data class.
The evaluation protocol is designed around standard computer vision datasets due to the limited availability of dedicated benchmarks~\cite{Hermary2025RemovingGB}.
We evaluate on CIFAR-10~\cite{Krizhevsky2009LearningML}, an image classification dataset with 10 semantic classes.
10 evaluation scenarios are constructed, each defined by selecting one class as normal while treating all remaining classes as anomalous.
This protocol induces a severe class imbalance favouring the positive side, as the resulting distribution places 1 class against 9 classes aggregated as outliers.

\subsubsection{Image Anomaly Detection (IAD)}
For IAD, we use MVTec AD~\cite{Bergmann2019MVTecA} and ViSA~\cite{Zou2022SPottheDifferenceSP}, both of which consist of industrial inspection images with defect-free and defective samples~\cite{Hicsonmez2026ASO}.
Evaluation is typically reported at two levels: (i) image-level detection, where the task is to classify an image as normal or defective, and (ii) pixel-level localisation, where the goal is to segment defective regions.
These two perspectives induce markedly different imbalance regimes: while defect images may be relatively more frequent at the image level, defective pixels remain extremely sparse within each image, leading to a significantly lower anomaly ratio at the pixel level.

The anomaly percentages for all datasets are summarised in \Cref{tab:datasets}.
Images from IAD datasets and corresponding masks are resized to $500 \times 500$ prior to pixel-wise evaluation.

\subsection{Metrics}
\label{subsec:metrics}

We study four evaluation metrics commonly used across the anomaly detection applications introduced previously. Two metrics are threshold-independent and evaluate the ranking quality of anomaly scores: Area Under the Receiver Operating Characteristic Curve (AUROC) and Area Under the Precision--Recall Curve (AUPR). The two others require a decision threshold to convert anomaly scores into binary predictions: the F$_1$-score and Matthews Correlation Coefficient (MCC).
Formal definitions of the metrics underlying components can be found in the supplementary materials.

\subsubsection{F$_{\textrm{\textit{1}}}$}
The F$_1$-score measures the harmonic mean between precision and recall, balancing the ability to detect anomalies while limiting false alarms.
\[
\textrm{F}_1 :=
\frac{2 \times TP}{2 \times TP + FP + FN}
\]

As it does not account for true negatives, F$_1$ inherently emphasises performance on the positive class, making it well suited for settings where the positive class is the minority.

\subsubsection{MCC}
The Matthews Correlation Coefficient (MCC) measures the correlation between predictions and ground truth, accounting for all entries of the confusion matrix.
\[
\mathrm{MCC} :=
\frac{
TP \times TN - FP \times FN
}{
\sqrt{
\begin{aligned}
(T&P + FP)\times(TP + FN)\, \\
&\times(TN + FP)\times(TN + FN)
\end{aligned}
}
}
\]

MCC is fully symmetric and is generally regarded as more robust to class imbalance than many standard classification metrics.
However, this symmetry can reduce interpretability, and its dependence on a decision threshold can lead to increased instability under extreme imbalance.
In this study, we report a scaled version normalised to the  [0,1] range.

\subsubsection{AUROC}
AUROC evaluates the ability of a model to rank anomalous samples ahead of normal ones across all possible thresholds.
It represents the probability that a randomly chosen positive instance will be ranked higher than a randomly chosen negative instance.
\[
\textrm{AUROC} := \int_0^1 TPR\left(FPR^{-1}\left(x\right)\right)dx
\]

Because AUROC is based on rates, it does not prioritise any class, even in imbalance cases; however, it does not reflect absolute error counts.
In highly imbalanced settings with many negatives, a large number of false positives can correspond to only a small increase in FPR, making the metric less sensitive to errors on the majority class in absolute terms.

\subsubsection{AUPR}
AUPR measures the trade-off between precision and recall across all thresholds, emphasising the quality of positive predictions.
\[
\textrm{AUPR} := \int_0^1 PPV\left(TPR^{-1}\left(x\right)\right)dx
\]

In settings dominated by the negative class, AUPR avoids dilution from true negatives by design, as it does not incorporate them.
Unlike AUROC, it is therefore highly sensitive to absolute changes in false positives and false negatives.
It effectively emphasises the positive class, and its value can decrease significantly as the proportion of positive samples becomes smaller.

\begin{figure}[t]
    \centering
    \includegraphics[width=\linewidth]{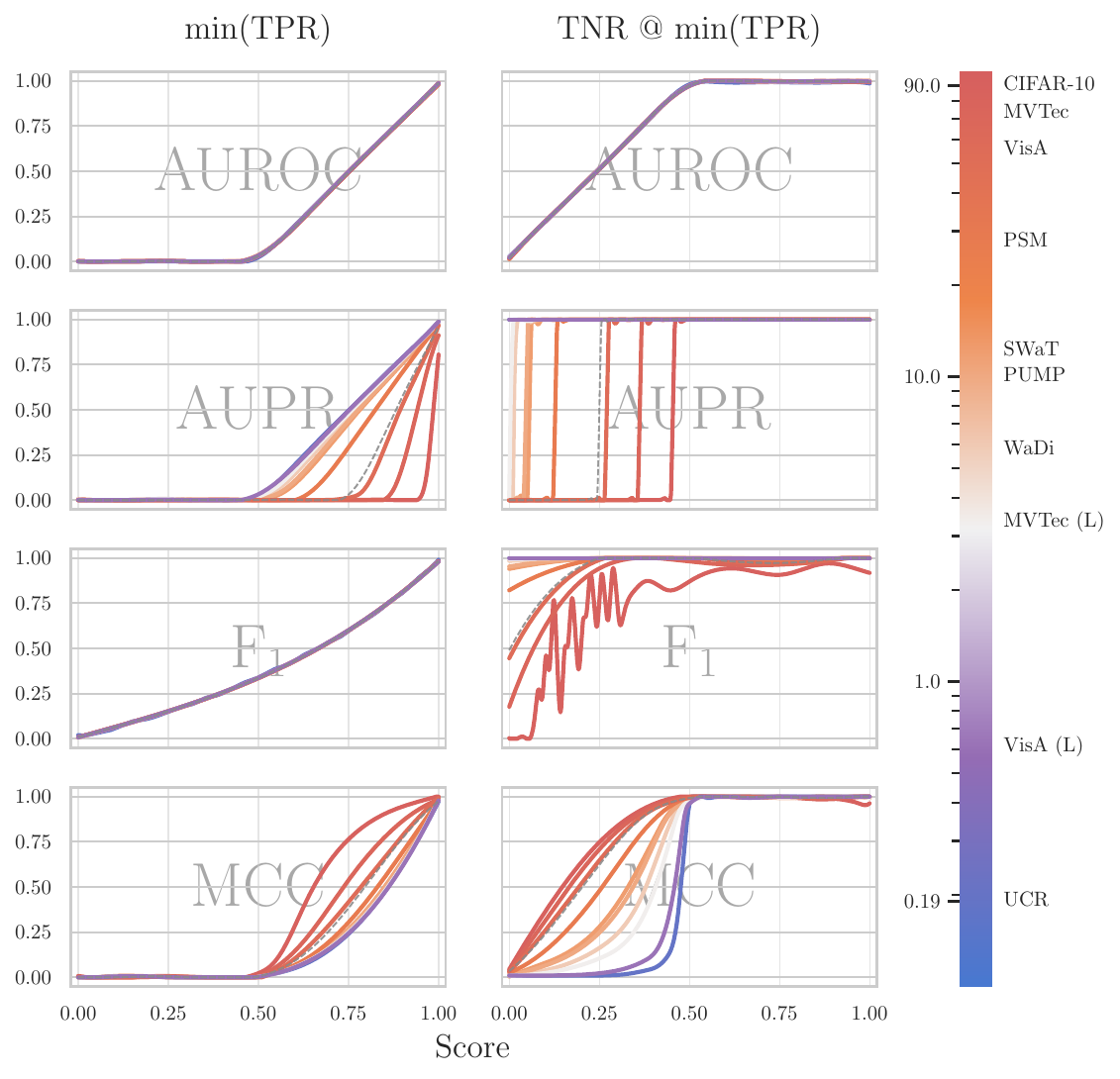}\\[-.5em]
    \caption{
\textit{Left:} Minimum TPR (y-axis) required to achieve the target score (x-axis). \textit{Right:} Corresponding TNR (y-axis) at minimum TPR required to achieve the score (x-axis).
Values at the different datasets imbalance ratios are shown in colour, while the balanced case is represented with dashed lines.
}
    \label{fig:mintpr}
\end{figure}

\section{Null Analysis}

We first establish a baseline for all metrics under random predictions.
For each imbalance ratio ranging from 99:1 to 1:99 (\textit{positive:negative)}, we sample 100 runs of random predictions and compute the mean, standard deviation, and mean positive deviation of each metric (\Cref{fig:stability}).

In the balanced case (50:50), all metrics centre around 0.5.
As the proportion of positive samples increases (left side of the plot), both AUPR and F$_1$ increase, raising the effective baseline to approximately 0.70 for F$_1$.
In extreme regimes where positives dominate, comparing AUPR results  becomes less informative, as high scores (\eg, 0.90 on CIFAR-10) can still correspond to random predictions.

In contrast, AUROC and MCC remain centred around 0.5 across all imbalance ratios.
However, AUROC exhibits higher variability and a mean positive deviation reaching approximately 0.25, indicating the likelihood of a single favourable run yielding deceptively strong scores (\eg, $\approx\,$0.75 AUROC).
AUPR also shows non-negligible upward deviations (up to $\approx\,$0.10 near the imbalance regime of MVTec), despite relatively stable standard deviation.

By comparison, F$_1$ and MCC remain concentrated, with standard deviation and mean positive deviation typically below 0.05, making them less prone to spurious high scores under random behaviour.

Finally, the AUPR baseline varies approximately linearly with the anomaly proportion, which makes deviations from random performance easy to quantify. Its relatively low variance further contributes to its usefulness as a stable and informative metric for anomaly detection.

\begin{figure}[t]
    \centering
    \includegraphics[width=1\linewidth]{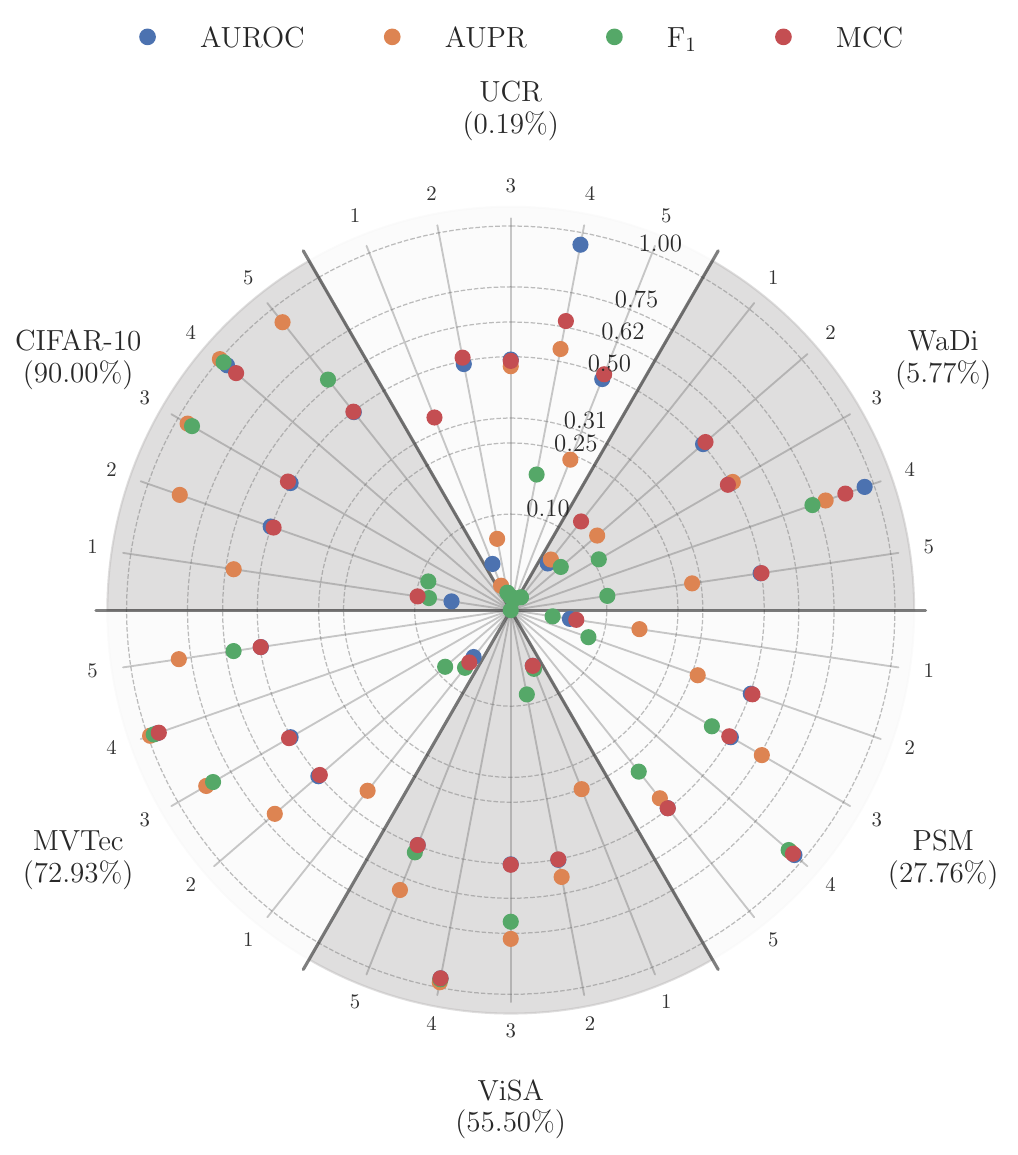}\\[-.5em]
\caption{
Average score within 5 different $10\%\times10\%$ (TPR$\times$TNR) metric landscapes regions (black boxes in \Cref{fig:landscapes}).
The evaluated landscapes are taken at the datasets ratios.
The radial scale is gamma-corrected ($\gamma=.6$).
}
    \label{fig:radar_regions}
\end{figure}

\section{Evolution Under Imbalance}

\subsection{Metrics Landscapes}

To gain an initial understanding of metric behaviour, we visualise score landscapes with respect to the TPR and TNR under the balanced setting (50:50) in \Cref{fig:landscapes}.

AUROC and MCC exhibit an approximately symmetric structure, with the baseline behaviour aligned along the diagonal $\left(\frac{\textrm{TPR} + \textrm{TNR}}{2} = 0.5\right)$.
In contrast, AUPR and F$_1$ place stronger emphasis on correct positive predictions, reaching higher values more rapidly along the TPR axis.
Additionally, F$_1$ stays low at low TPR values even with high TNR.

\textit{The five numbered regions shown on the AUROC landscape correspond to the cases analysed in detail in \Cref{subsec:regions}.}

\subsection{Global Evolution}

To analyse how score distributions evolve under class imbalance, we study the difference between landscapes at different imbalance ratios.
As the evolution appears approximately linear, we focus on a representative case at 5:95.
We examine the global difference between the balanced setting and the negative-majority regime (\Cref{fig:evol-neg}).
The symmetric positive-majority case is provided in the supplementary material.

As expected, AUROC shows no change, reflecting its invariance to class imbalance.
In contrast, the other metrics exhibit substantial variations.
For AUPR, most of the landscape decreases by at least 0.2, with reductions reaching up to 0.4 in regions of increasing TNR.
Only narrow regions with simultaneously high TPR and TNR remain relatively stable.
This reflects the reduced influence of negative-class performance as imbalance increases.

F$_1$ is also strongly affected by imbalance, although true negatives retain more influence than for AUPR.
Even with perfect TPR, F$_1$ only reaches approximately 0.5 when TNR is around 0.6 (landscape colour).
However, given that the baseline F$_1$ at this ratio is below 0.1 (\Cref{fig:stability}), even modest absolute values correspond to substantial improvements over random performance.

MCC exhibits a more complex pattern.
Extreme TNR values (around 0 and 1) remain relatively stable, while values along the diagonal stay near 0.5.
Below the diagonal, scores increase, whereas above it they decrease, compressing the distribution around mid-range values and making average score interpretation less direct.
Regions with minimal TNR decrease, while those with maximal TPR increase sharply, indicating a stronger dependence on the negative class under imbalance.

\begin{figure}[t]
    \centering
    \includegraphics[width=.95\linewidth]{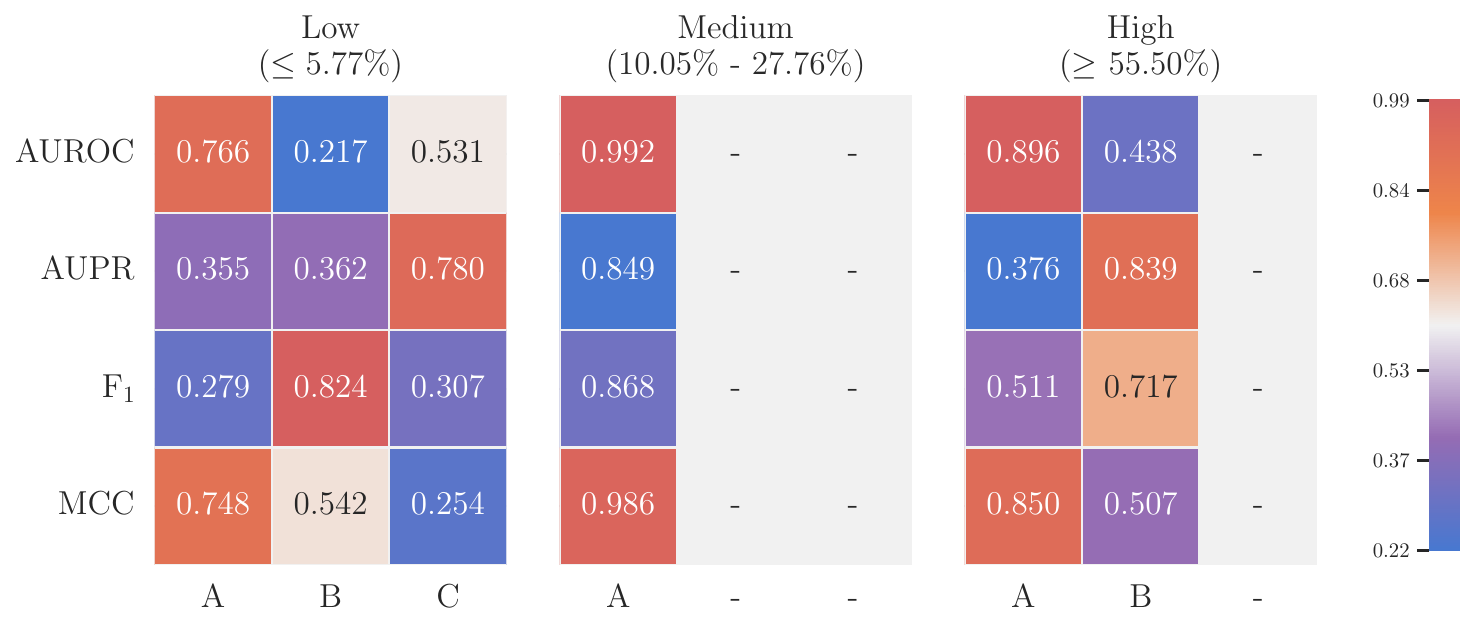}\\[-.5em]
    \caption{
Factor analysis of the 4 metrics over the studied datasets, grouped into 3 categories (Low, Medium, High) based on their anomaly percentages.
Orthogonal variance maximisation rotation is applied and factors signs aligned.
}
\label{fig:factor_analysis}
\end{figure}

\subsection{Minimal Positives Detection}

Analysing global patterns, we also determined the minimal ratios of detected positives for different scores and imbalance ratios~(\Cref{fig:mintpr}, left column).
Additionally, we computed in the right column the corresponding TNR required to achieve said score at the identified minimum TPR.

For AUROC, no positive sample needs to be detected up to a score of 0.5.
Beyond this point, assuming all negatives are correctly ranked, the score increases progressively as more positives are correctly ranked.

For AUPR, no positive detection is required up to a score of 0.75 in a balanced dataset.
This threshold shifts \wrt{} the anomaly ratio; however, achieving such scores requires 100\% of the negatives to be correctly classified.

F$_1$ exhibits a particularly interesting behaviour.
Regardless of the anomaly ratio, increasing the score always requires an increasing number of correctly classified positives.
However, as the anomaly ratio decreases, a larger number of correctly classified negatives is also required to reach the same score.

Finally, MCC displays a pattern similar to AUPR, although smoother and reversed \wrt{} the anomaly ratio.
Up to a score of 0.5, no correctly classified positive is required.
Concurrently, the proportion of correctly classified negatives needed decreases more rapidly as the anomaly ratio decreases.
The extreme case of UCR shows that, on highly imbalanced datasets, achieving an MCC of $\approx\,$0.4 is possible without detecting any positives and with barely 10\% of negatives correctly classified.
Nevertheless, this value remains below the random MCC baseline~(\Cref{fig:stability}).

\begin{figure}[t]
    \centering
    \includegraphics[width=\linewidth]{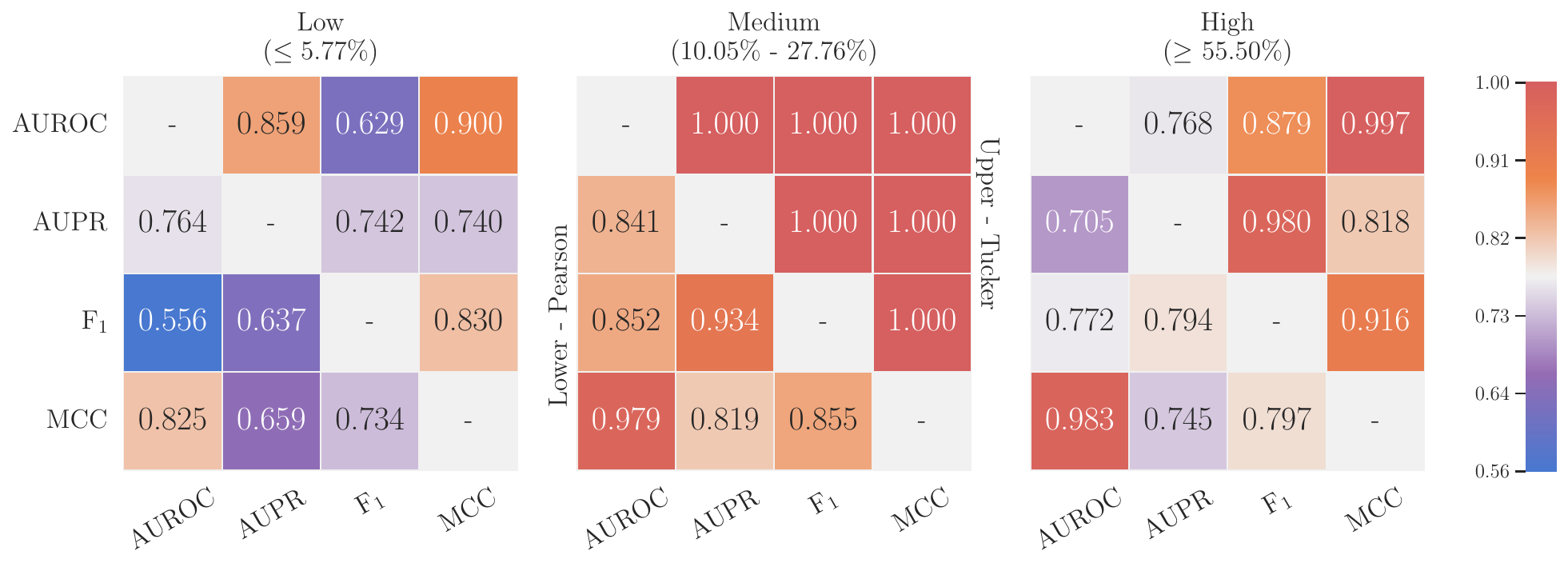}\\[-.5em]
    \caption{
\textit{Upper triangle:} Tucker's congruence coefficients between metrics factor loadings (\Cref{fig:factor_analysis}).
\textit{Lower triangle:} Person's correlation coefficients between the metric landscapes.
}
    \label{fig:tucker}
\end{figure}

\subsection{Regions of Interest Analysis}
\label{subsec:regions}

In \Cref{fig:radar_regions}, we report on several edge-case regions of the metrics landscapes, corresponding to characteristic prediction patterns: predicting almost everything incorrectly (R1); almost all negatives (R2); almost all positives (R3); almost everything correctly (R4); and approximately half correctly (R5).

\textit{(R1) TPR 0-10\%; TNR 0-10\%}.
All metrics remain very low regardless of the anomaly ratio.
MCC shows a slight increase for the highly imbalanced UCR and WaDi cases.
In contrast, AUPR increases with the anomaly percentage, reaching approximately 0.5 for CIFAR-10.

\textit{(R2) TPR 0-10\%; TNR 90-100\%}.
AUROC and MCC consistently remain around 0.5 across all anomaly ratios.
AUPR again increases with the anomaly ratio, closely following its random baseline.
F$_1$ remains very low for all imbalance settings, making it comparatively stable in the edge case where nearly all predictions are negative.

\textit{(R3) TPR 90-100\%; TNR 0-10\%}.
In this case, both MCC and AUROC produce the same characteristic score of $\approx\,$0.5, independently of the anomaly ratio.
Conversely, F$_1$ and AUPR increase with the anomaly percentage and can approach 1 in the most balanced cases.

\textit{(R4) TPR 90-100\%; TNR 90-100\%}.
All metrics naturally approach 1.
However, for highly imbalanced datasets such as WaDi, AUPR, MCC, and F$_1$ exhibit a slight decrease in their average values within this region, highlighting a sharp contribution of the high TNR and TPR to the score.

\textit{(R5) TPR 45-55\%; TNR 45-55\%}.
AUROC and MCC remain stably centred around 0.
AUPR increases with the anomaly ratio, while F$_1$ exhibits the strongest sensitivity to class imbalance, ranging from nearly 0 in the UCR setting to approximately 0.65 in the most balanced case.
This region unsurprisingly closely matches the behaviour expected from random baseline predictions.

\section{Metrics Relationships}

Factor analysis was performed to investigate the latent relationships between evaluation metrics and analyse how they vary across different anomaly ratio regimes.
The study is conducted separately for three anomaly ratio groups: Low ($\leq5.77\%$: UCR, VisA (L), MVTec (L), and WaDi), Medium (10.05\%-27.76\%: PUMP, SWaT, and PSM), and High ($\geq55.50\%$: VisA, MVTec, and CIFAR-10).

The loadings of the extracted factors (A, B, and C), selected to retain 90\% of the explained variance, are shown in~\Cref{fig:factor_analysis}.
The corresponding Tucker congruence coefficients between metrics are presented in~\Cref{fig:tucker} (upper triangle), alongside the Pearson correlations computed directly from the score landscapes (lower triangle).
Pearson correlation reflects local similarity in the observed score distributions, whereas Tucker’s congruence captures similarity in the latent factor structure shared by the metrics.

In the Medium regime, all metrics exhibit similar loading profiles, indicating that they capture closely related structures.
This is further supported by the high Pearson correlations observed across all metric pairs, suggesting strong agreement under moderate imbalance conditions.

For highly imbalanced datasets dominated by negative samples (Low), the metrics become more complementary, as reflected by the decrease in both Tucker congruence and Pearson correlation values.
This indicates that the metrics no longer respond similarly across the score landscapes and emphasise different regions or classification behaviours.
AUROC and MCC remain the most closely related metrics, whereas F$_1$ and AUROC display the weakest agreement, both in the latent structure and in the raw landscape correlations.

In the High regime, the metrics exhibit increased decorrelation compared to the Medium regime, although the effect is generally less pronounced than in the Low regime.
The pair AUPR-AUROC notably shows reduced congruence and correlation, indicating increasingly different behaviours as the positive class becomes dominant.
In contrast, AUROC and MCC maintain total similarity, with a Tucker congruence of 0.997 and a Pearson correlation of 0.983, demonstrating that both metrics retain nearly identical structural behaviour across the score landscapes despite changes in the anomaly ratio.
\section{Conclusion}

common anomaly detection evaluation metrics evolve under varying anomaly ratios.
Through the study of metric landscapes, based on true positive and true negative rates, with insightful experiments, we provided practical guidance for interpreting the metrics under specific class imbalances and different points of view.
Future work will extend this analysis to other anomaly detection domains and their datasets characteristics, such as in industrial monitoring, internet of things, cybersecurity, or autonomous driving.

\section*{Acknowledgment}

This research was funded by the Luxembourg National Research Fund (FNR), grant reference \texttt{DEFENCE22/17813724/AUREA}.

\bibliographystyle{IEEEtran}
\bibliography{bib2}

\clearpage
\appendices


\section{Details on Metrics Elements}

\Cref{tab:elements} Summarises the different components of the metrics formulas discussed in the main paper.

\begin{table}[h]
    \centering
    \caption{Metrics components.}
    \resizebox{\linewidth}{!}{
\renewcommand{\arraystretch}{1.15}%
\begin{tabular}{ll}
\toprule
Quantity & Formula \\
\midrule
True Positives (TP) &
$\mathrm{TP} = \sum_i \mathds{1}(\hat{y}_i = 1 \land y_i = 1)$ \\

True Negatives (TN) &
$\mathrm{TN} = \sum_i \mathds{1}(\hat{y}_i = 0 \land y_i = 0)$ \\

False Positives (FP) &
$\mathrm{FP} = \sum_i \mathds{1}(\hat{y}_i = 1 \land y_i = 0)$ \\

False Negatives (FN) &
$\mathrm{FN} = \sum_i \mathds{1}(\hat{y}_i = 0 \land y_i = 1)$ \\

True Positive Rate (TPR / Recall) &
$\mathrm{TPR} = \frac{\mathrm{TP}}{\mathrm{TP}+\mathrm{FN}}$ \\

False Positive Rate (FPR) &
$\mathrm{FPR} = \frac{\mathrm{FP}}{\mathrm{FP}+\mathrm{TN}}$ \\

Positive Predictive Value (PPV / Precision) &
$\mathrm{PPV} = \frac{\mathrm{TP}}{\mathrm{TP}+\mathrm{FP}}$ \\
\bottomrule
\end{tabular}

}
    \label{tab:elements}
\end{table}

\section{Score Coverage}

\begin{figure}[t]
    \centering
    \includegraphics[width=1\linewidth]{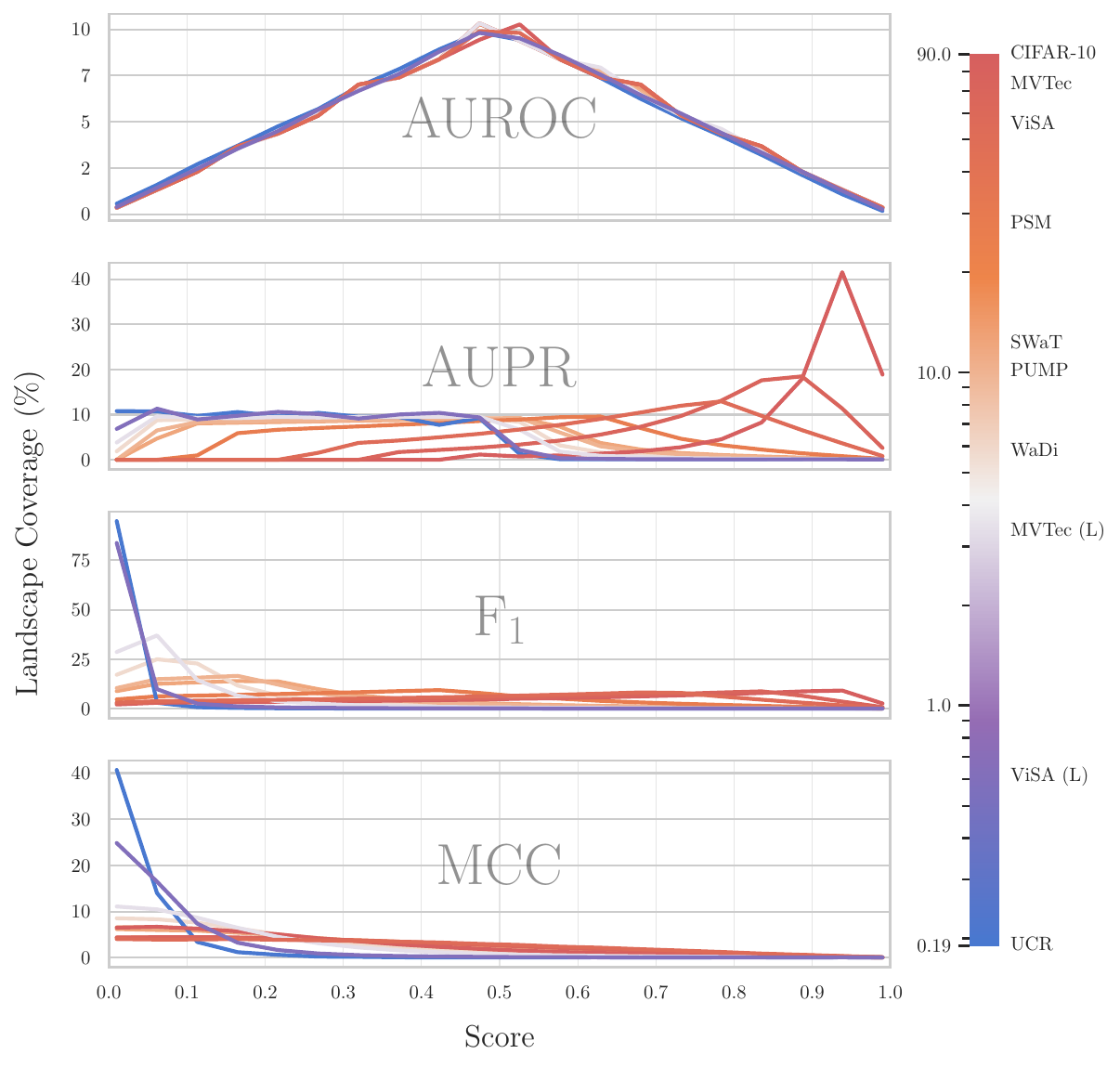}\\[-.5em]
    \caption{
Proportion of surface area (\%) from landscape corresponding to each score value for different anomaly ratios (line colour).
Scores were discretised into 20 non-overlapping bins (width of .05 points) to mitigate sampling noise.
}
    \label{fig:coverage}
\end{figure}

In order to understand the scores distributions at different imbalance level, we provide a surface coverage analysis in \Cref{fig:coverage}.

Concretely, the surface area equals to each score was computed \wrt{} the total landscape area, giving a coverage percentage.
To smooth the process, 20 bins or ranges were used instead of specific scores, so that a line point corresponds to a score $\pm0.05$.

For AUROC, we observe that most of the area is covered by the average score of 0.5.
Combining those results with the landscapes visualisation, both support the idea of linearly increasing difficulty of getting to a higher score, with higher average of ratios $\frac{\textrm{TPR} + \textrm{TNR}}{2}$.

AUPR has a relatively stable score distribution with low anomaly percentages.
Most scores below 0.5-0.6 have a coverage of 10\%, meaning that they these bring a relative uncertainty as to how the model actually performs, \wrt{} TPR and TNR.
High anomaly percentage has a worse pattern of uncertainty, with more than 40\% of the landscape corresponding to scores of $\approx\,$0.95 for CIFAR-10 imbalance ratio.
This should raise awareness on the unreliability of AUPR in such cases.

F$_1$ and MCC have similar patterns, and their low coverages throughout all scores should make them great metrics to easily pinpoint the model correct predictions distributions.
We can observe in highly negative-dominant datasets, like UCR and ViSA (L), a spike in $\approx\,$0 score coverage, emphasising the increasing difficulty in getting higher scores.

\section{Landscape Evolution: Positive Majority}

\begin{figure*}[t]
    \centering
    \includegraphics[width=1\linewidth]{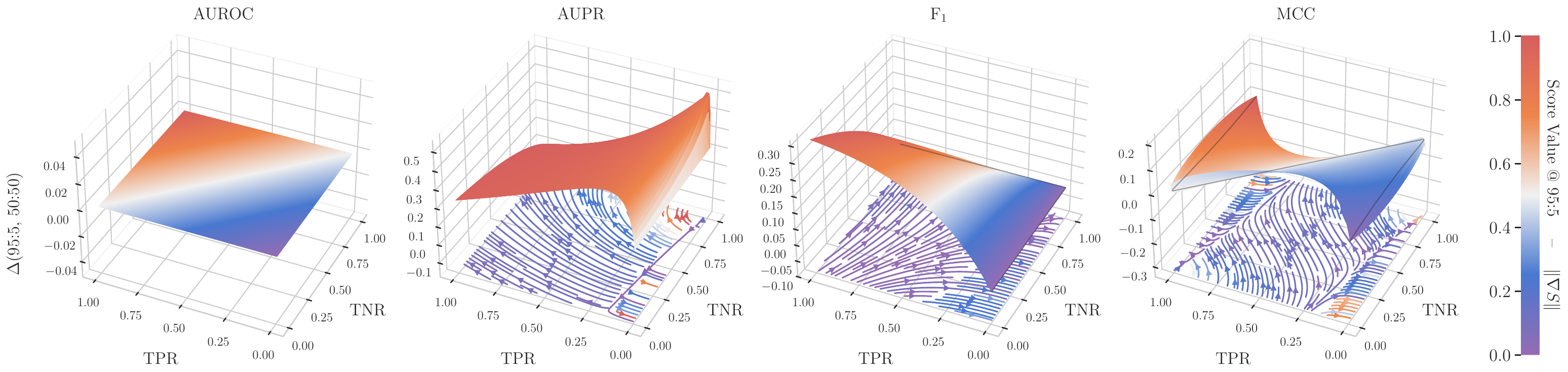}\\[-.5em]
    \caption{
Landscape evolution (difference) between balanced and positive majority cases.
$\Delta(a, b) = a-b$.
}
\label{fig:evol-pos}
\end{figure*}

\Cref{fig:evol-pos} presents the landscape evolution study upon positive majority.
As for the negative-majority case, and without surprise, AUROC is flat and insensitive to class imbalance.

\begin{figure*}[t]
    \centering
    \includegraphics[width=1\linewidth]{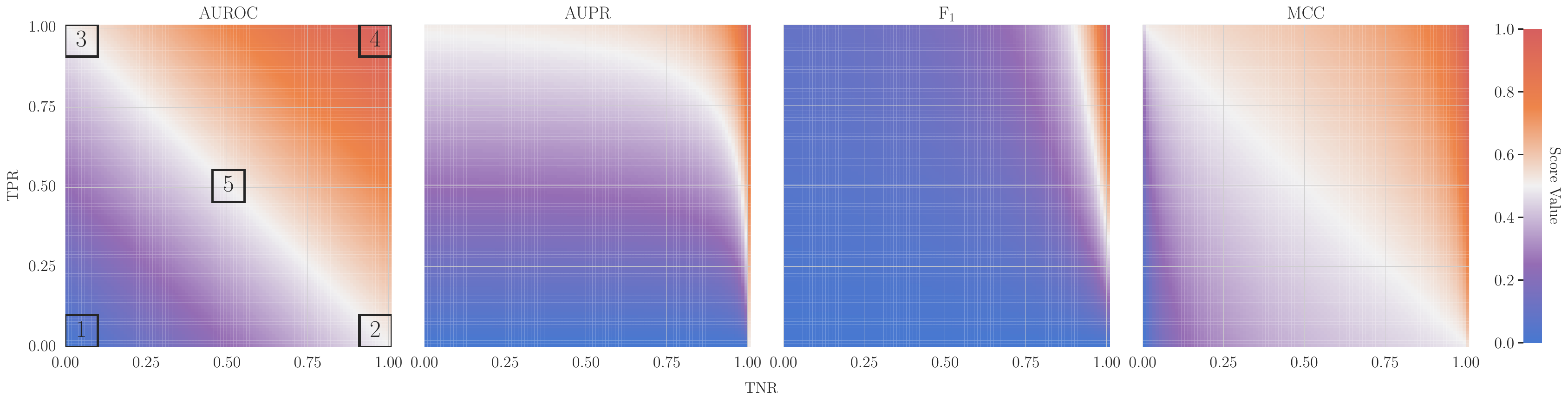}\\[-.5em]
    \caption{
Metrics landscapes in the negative majority case (5:95).
}
    \label{fig:landscapes5}
\end{figure*}
AUPR shows up to almost 0.5 of increase score value compared to the balanced case at the same TP and TN rates.
The increase is particularly present at low TPR, but the sharp edge and decreasing slope between 0.0 and 0.25 TPR, combined the surface colour (actual scores values), suggest that the highest scores ($>0.6$) will only be achieved by predicting the positives correctly. 

F$_1$ behaves in the almost exact opposite, with highly increased scores when predicting positives correctly.
In this case, the majority class is favoured and superior scores compared to a balanced case will be achieved easily.

MCC has again the most complex evolution, although with a similar symmetric pattern compared to the negative-majority case.
Values around the average baseline score of 0.5 will also be increasingly present on the landscape.

\begin{figure*}[h]
    \centering
    \includegraphics[width=1\linewidth]{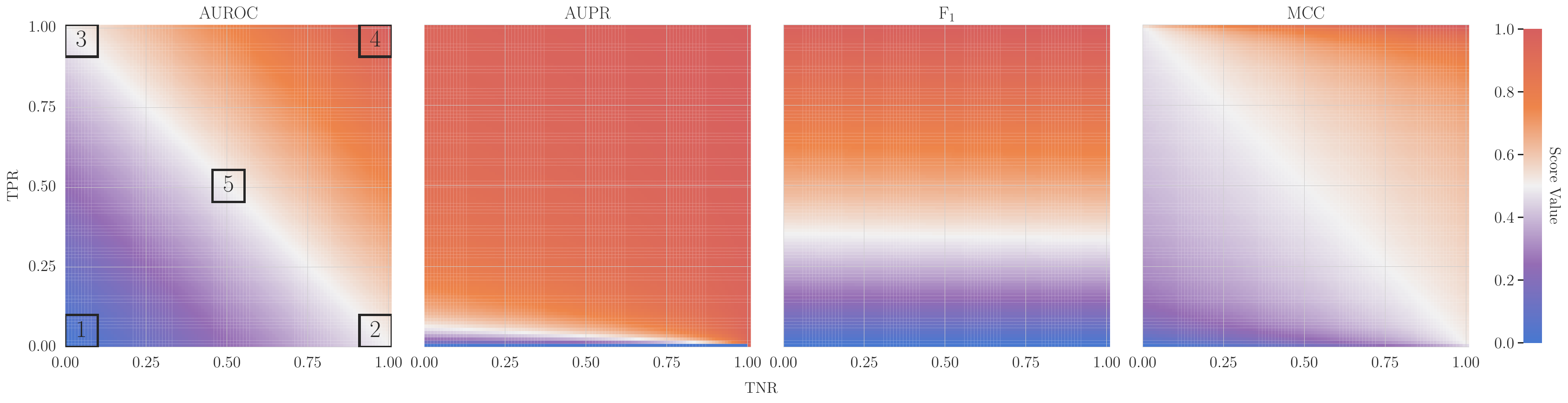}\\[-.5em]
    \caption{
Metrics landscapes in the negative majority case (95:5).
}
    \label{fig:landscapes95}
\end{figure*}

\begin{figure*}[h]
    \centering
    \includegraphics[width=1\linewidth]{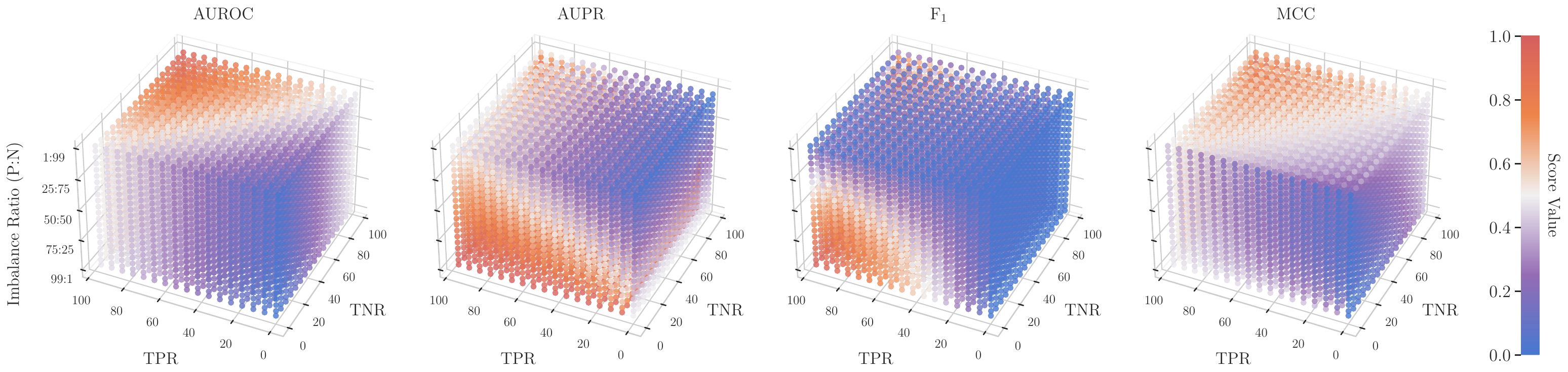}\\[-.5em]
    \caption{Landscapes from different ratios superposed on z-axis. Downsampling was applied and sub-regions replaced with dots to improve visibility.}
    \label{fig:volumes}
\end{figure*}

The metrics landscapes at 95:5 and 5:95 imbalance ratios (P:N) are displayed plainly in \Cref{fig:landscapes95} and \Cref{fig:landscapes5}, respectively, and an overall view of the landscapes evolution over all ratios can be found under a volume representation in \Cref{fig:volumes}.

\section{Regions of Interest: Complete Study}

\begin{figure}[h]
    \centering
    \includegraphics[width=1\linewidth]{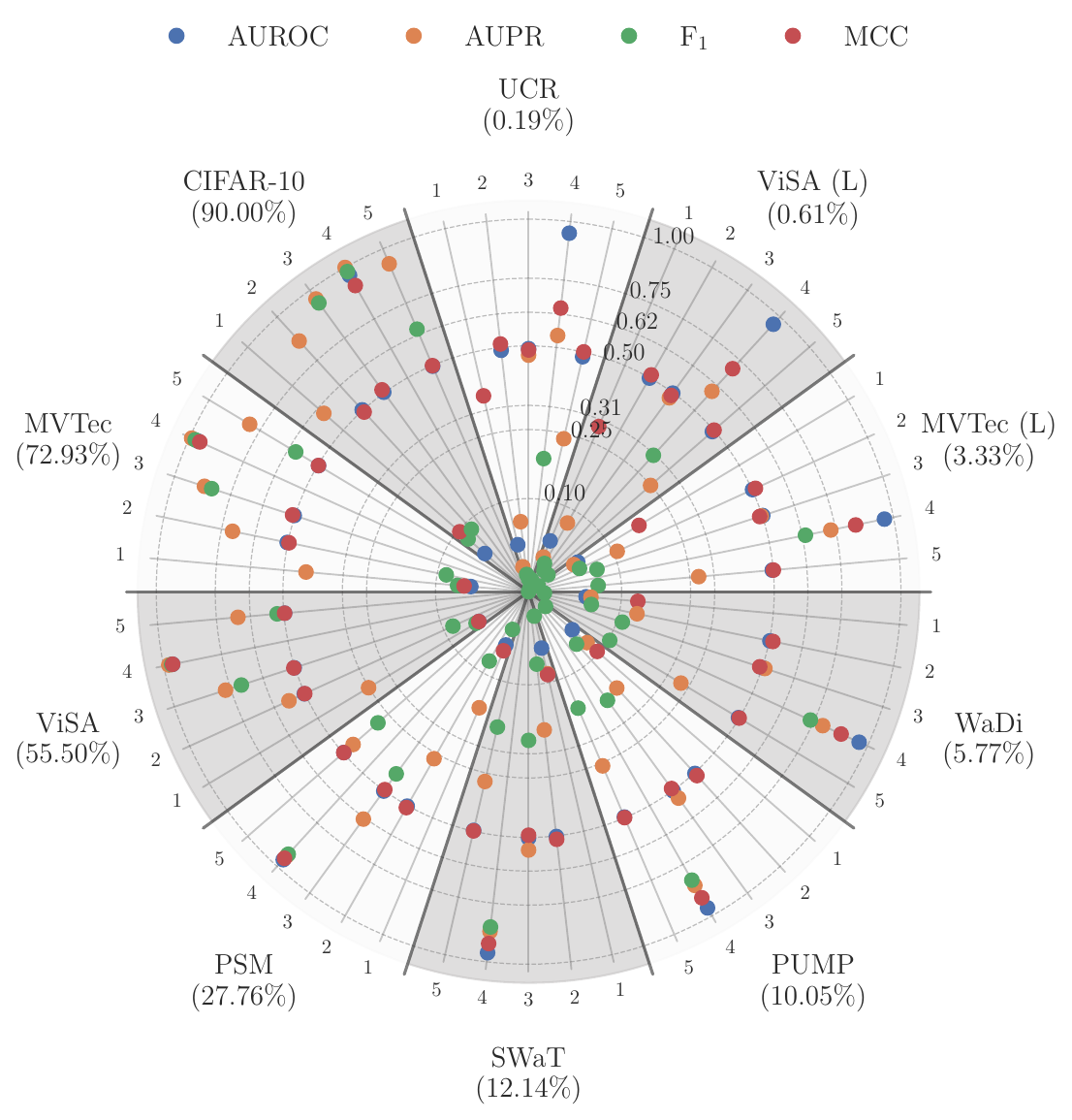}\\[-.5em]
    \caption{
Average score within 5 different $10\%\times10\%$ (TPR$\times$TNR) metric landscapes regions.
The evaluated landscapes are taken at the datasets ratios.
The radial scale is gamma-corrected ($\gamma=.6$).
}
    \label{fig:radar-all}
\end{figure}

\begin{figure}[h]
    \centering
    \includegraphics[width=1\linewidth]{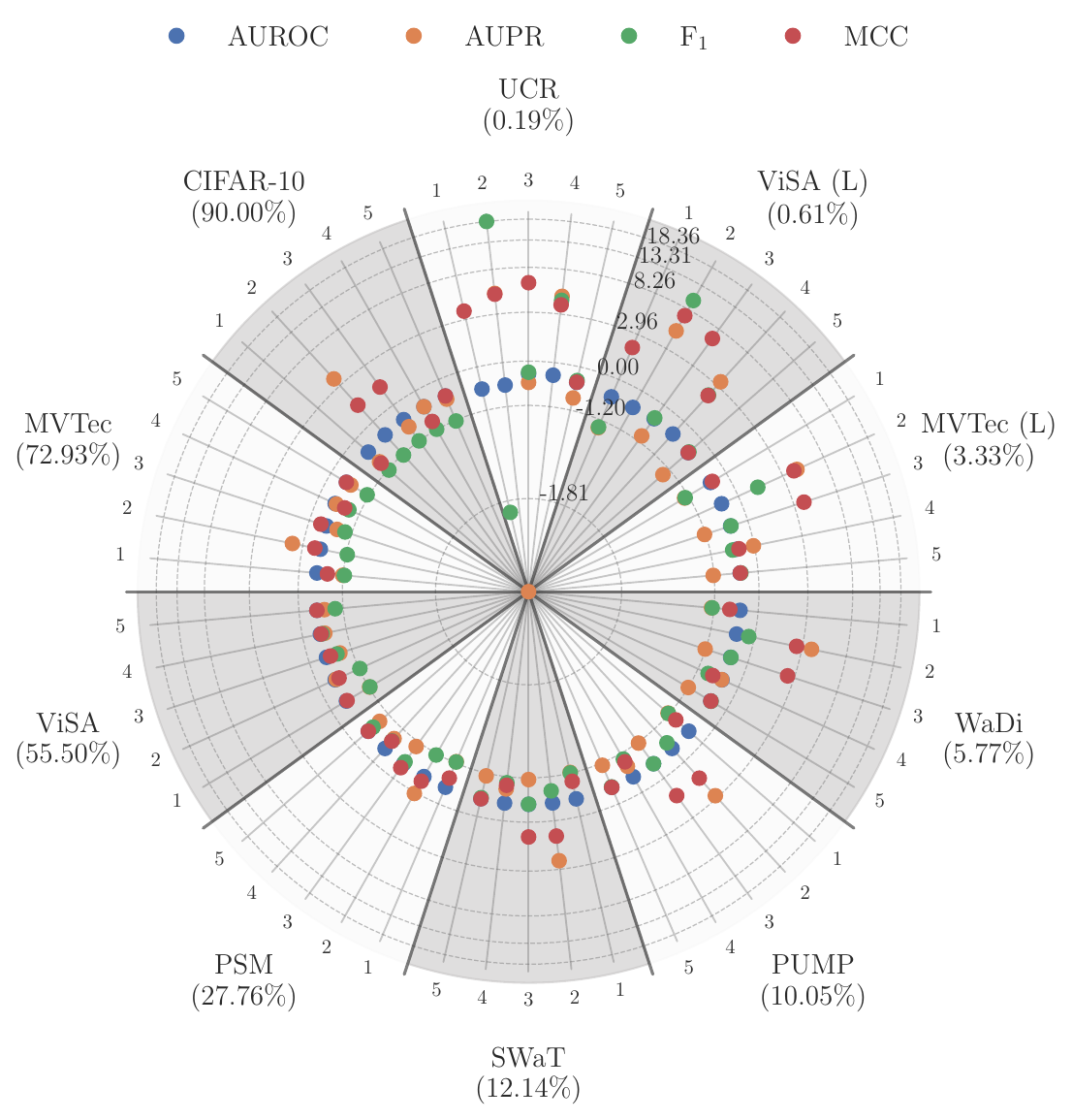}\\[-.5em]
    \caption{
Excess kurtosis within 5 different $10\%\times10\%$ (TPR$\times$TNR) metric landscapes regions.
The evaluated landscapes are taken at the datasets ratios.
The radial scale is gamma-corrected ($\gamma=.2$).}
    \label{fig:kurtosis-all}
\end{figure}

The full plot (all datasets) of the mean values in the five regions of interest is presented in \Cref{fig:radar-all}.
It is supplemented with the study of excess kurtosis values in those regions (\Cref{fig:kurtosis-all}).
We observe rather homogeneous structures, with mostly negative excesses.
MCC has the most often heavy-tailed distributions in those regions, with positive excess kurtosis values peaking at almost 8.
This showcases sharp changes in values in the studied regions.

\end{document}